\pdfoutput=1

\documentclass[11pt]{article}

\usepackage{acl}

\usepackage{times}
\usepackage{latexsym} 

\usepackage[T1]{fontenc}

\usepackage[utf8]{inputenc}

\usepackage{microtype}

\usepackage{inconsolata}

\usepackage{graphicx}
\usepackage{float}
\usepackage{subfigure}
\usepackage{multirow}
\usepackage{amsmath}
\usepackage{algorithmic,algorithm}
\usepackage{tcolorbox}
\usepackage{adjustbox}
\usepackage{cleveref}
\usepackage{enumitem}

\usepackage{soul}

\title{Consultant Decoding: Yet Another Synergistic Mechanism}

\author{
 \textbf{Chuanghao Ding\textsuperscript{1}},
 \textbf{Jiaping Wang\textsuperscript{2}},
 \textbf{Ziqing Yang\textsuperscript{3}},\\
 \textbf{Xiaoliang Wang\textsuperscript{1}},
 \textbf{Dahua Lin\textsuperscript{4}},
 \textbf{Cam-Tu Nguyen\textsuperscript{1}}\thanks{Corresponding authors},
 \textbf{Fei Tan\textsuperscript{2}}\footnotemark[1]
\\
\\
 \textsuperscript{1}State Key Laboratory for Novel Software Technology, Nanjing University,\\
 \textsuperscript{2}East China Normal University,
 \textsuperscript{3}Independent Researcher,
 \textsuperscript{4}The Chinese University of Hong Kong
\\
 \small{
 ch777.ding@smail.nju.edu.cn\ \ \ 
 51265902031@stu.ecnu.edu.cn\ \ \  
 \{waxili, ncamtu\}@nju.edu.cn
 } \\
 \small{
  ziqingyang@gmail.com\ \ \ 
  dhlin@ie.cuhk.edu.hk\ \ \ 
  tanfei2007@gmail.com
 }
}

\begin{document}
\maketitle

\begin{abstract}
The synergistic mechanism based on Speculative Decoding (SD) has garnered considerable attention as a simple yet effective approach for accelerating the inference of large language models (LLMs). 
Nonetheless, the high rejection rates require repeated LLMs calls to validate draft tokens, undermining the overall efficiency gain of SD.
In this work, we revisit existing verification mechanisms and 
propose a novel synergetic mechanism \textit{Consultant Decoding} (CD). 
Unlike SD, which relies on a metric derived from importance sampling for verification, CD verifies candidate drafts using token-level likelihoods computed solely by the LLM.
CD achieves up to a 2.5-fold increase in inference speed compared to the target model, while maintaining comparable generation quality ($\sim100\%$ of the target model’s performance). Interestingly, this is achieved by combining models whose parameter sizes differ by two orders of magnitude.
In addition, CD reduces the call frequency of the large target model to below $10\%$, particularly in more demanding tasks.
CD's performance was even found to surpass that of the large target model, which theoretically represents the upper bound for speculative decoding.

\end{abstract}

\section{Introduction}
\label{sec:intro}

Large Language Models (LLMs) have attracted widespread attention and adoption across diverse domains, owing to their exceptional performance and robust generalization abilities \cite{yang2024qwen25, lu2023makes, wang2024reward, zhang2024balancing, gu2024cmr, li2023cwseg, dubey2024llama3}. As their capabilities continue to evolve, so does their scale, with some reaching unprecedented sizes. Notable examples include LLaMA3.1-405B \cite{dubey2024llama3} and DeepSeek-V3/R1 models \cite{liu2024deepseek}. The growing size of large models is making inference efficiency an increasingly pressing concern.

\begin{figure}[t]
  \centering
  \subfigure[Speculative Decoding]{
        \centering
        \includegraphics[height=0.85in]{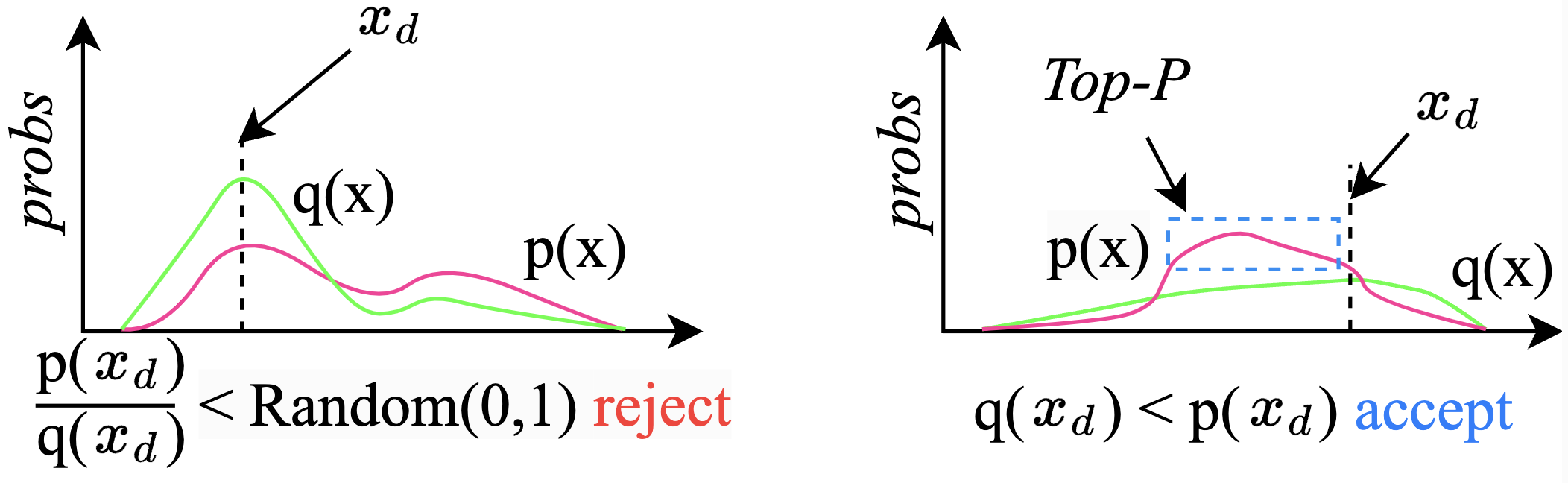}
        \label{fig:problem_SD}
    }
    \hfill
    \subfigure[Consultant Decoding]{
        \centering
        \includegraphics[height=0.9in]{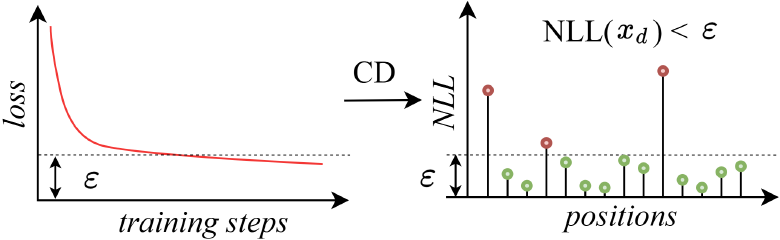}
        \label{fig:verify_CD}
    }
  \caption{(a) A visual description of the sub-optimal problem of distribution-based method, where the blue box is high-quality nucleus of Top-P sampling and the $x_d$ is the draft token. (b) The CD verification algorithm. $q(x)$ and $p(x)$ denote distribution of draft model and target model, respectively. The $\varepsilon$ is the approximate convergence loss in the training phase of target model.}
  \label{fig:verify_visualization}
\end{figure}

Recently, Speculative Decoding (SD)~\cite{speculative_google, speculative_deepmind}, based on the draft-and-verify paradigm, has been introduced to accelerate the decoding process of large language models (LLMs). 
It exploits the key property that parallel processing can improve GPU utilization and reduce the average latency during autoregressive decoding. SD maintains distribution consistency between the sampling results and target models.
This advantage has spurred the development of numerous improved methods~\cite{li2024eagle, li2024eagle2, zafrir2024fastdraft, monte_carlo_SD, faster_cascades, yang2024multi, miao2024specinfer, xiao2024clover2}.

However, these methods rely on a rigid verification process akin to speculative sampling, leading to an excessive rejection of draft tokens and frequent target model calls. These challenges raise a crucial question: \textit{Can we devise a better verification approach that boosts the acceptance rate and lessens reliance on target models, while maintaining a high level of performance?}

In this paper, we examine existing speculative decoding verification approaches~\cite{speculative_google, mentored_decoder} and observe that they predominantly evaluate the draft tokens by likelihood ratio or its variants between the target and draft models, as illustrated in \Cref{fig:problem_SD}. This metric is derived from SD's importance sampling to ensure that the final sampling distribution is well aligned with that of the LLM.
However, the likelihood ratio does not reflect well the quality of the generated token. For example, as shown in Figure~\ref{fig:problem_SD}, when the target model's prediction probability marginally exceeds that of the draft model – even if both probabilities are substantially low – the candidate tokens will still be accepted, potentially compromising output quality.
In addition, the target model’s best token may still be rejected, particularly when the target language model has a relatively flat probability distribution.
The enhanced SD approaches, such as Mentored Decoding (MD) ~\cite{mentored_decoder}, which continue to exhibit suboptimal problems, as it relies on likelihood ratio. Furthermore, it is unclear on how to choose the parameters, such as threshold and tolerance.

To address these problems, we introduce a novel approach that directly evaluates the correctness of candidate tokens using likelihood of the target model. Essentially, we leverage the negative log-likelihood (NLL) of candidate tokens predicted by target models to directly evaluate their correctness, a method we refer to as \textit{Consultant Decoding} (CD).
As illustrated in Figure~\ref{fig:verify_CD}, a draft token is deemed correct if its NLL approaches or falls below the convergence loss observed during the training of target models, signaling that the distribution of draft tokens closely mirrors the training data distribution of targe models. Consultant Decoding is named for its approach of consulting the target model's judgment to validate draft tokens, similar to seeking expert advice while retaining the ability to make independent decisions. This approach strikes a balance between independence (the draft model's autonomy) and collaboration (the target model's validation), leading to faster inference without sacrificing accuracy much even exceeding target models.

We conduct extensive empirical experiments using both publicly available datasets (covering areas such as math, coding, conversation, and instruction following) and real-world industrial scenarios, and observe significant benefits. 
Additionally, we compare the consistency of CD and Top-P sampling and found theoretical evidence that CD can maintain the performance of target models. Our analysis of speedup ratios and LLM-calls ratios also revealed that CD shows better scalability.

\section{Related Work}
\label{related-work}
\label{speculative-decoding}
Speculative decoding~\cite{speculative_pecking, speculative_google, speculative_deepmind} has been widely adopted and extended across various works, yielding notable speedups in numerous model families and datasets. Several strategies for optimizing the draft model to accelerate the generation of draft tokens have also been explored in the literature. These approaches include training a smaller draft model specifically optimized for faster token generation~\cite{cai2401medusa, li2024eagle, li2024eagle2, zafrir2024fastdraft, xiao2024clover, xiao2024clover2} and utilizing non-causal algorithms to generate draft tokens~\cite{fu2024break, zhao2024ouroboros, liu2024sdsat}, which enable parallelization of the token generation process, reasonably boosting throughput.

Additionally, several research efforts have focused on improving the acceptance rate of draft tokens by relaxing the verification criteria~\cite{monte_carlo_SD, faster_cascades, big_little_decoder}. Instead of strictly requiring that the distribution from speculative sampling aligns perfectly with the distribution of LLMs, these approaches allow for a certain degree of deviation within an acceptable range.  
These methods achieve a speedup by moderately sacrificing predictive accuracy.

Research efforts have also been undertaken to examine the structure of the drafts with the aim of boosting the acceptance rate of draft tokens: 1) Multi-candidate draft: These methods~\cite{yang2024multi, miao2024specinfer, wang2024opt_tree, brown2024dynamic, cai2401medusa, li2024eagle} involve using tree-based attention mechanisms to generate multiple candidate draft token sequences during each iteration for chances of accepting more tokens. 2) Block-Wise Verify: Researchers have also explored moving beyond the token-level verification of SD, opting instead for block-wise conditional probabilities ~\cite{monte_carlo_SD, sun2024block}. 
3) Dynamic Draft Length: In an effort to minimize the rejection of draft tokens for improving the acceptance rate~\cite{mamou2024disco, huang2024specdec++}. 
Existing works, such as BiLD~\cite{big_little_decoder}, have employed similar verification methods to ours by evaluating token accuracy through the distribution distance between the hard label of the draft model and the soft label of the target model.

Our work focuses on the novel non-distribution verification mechanism, which differs remarkably from the above diverse research threads.

\section{Method}
\label{method}
\subsection{Preliminaries}
\label{sec:preliminary}
\paragraph{Speculative Sampling.}
Speculative sampling typically requires two models: a large target model $P$ and a small draft model $Q$. The process involves iterative drafting and verification.

During the drafting phase, in each iteration, given the input token sequence $x_{\leq t}$, the draft model generates $\gamma$ candidate tokens in an autoregressive manner, denoted as $\{x_i, q_i(\cdot) \mid 1 \leq i \leq \gamma \}$, here $x_i$ represents the token sampled by the draft model at position $t+i$, and $q_i(\cdot)$ indicates the corresponding probability distribution predicted by the draft model. The sequence of input tokens $x_{\leq t}$ is concatenated with the draft tokens $\{x_i \mid 1 \leq i \leq \gamma \}$ to form $\{x_{\leq t}, x_1, \ldots, x_{\gamma}\}$, which is then passed to the target model for the parallel verification.

During the verification phase, the target model generates the probability distributions $\{p_i(\cdot) \mid 1 \leq i \leq \gamma + 1\}$ for each draft token. Then a verification method $V_{\mathrm{SD}}(x_i)$ is invoked to decide whether to accept the draft token or resample a new token
\begin{align}
    &V_{\mathrm{SD}}(x_i) = r(x_i) - \mathrm{Rand}(0,1) \\
    &r(x_i) = \min(1, p_i(x_i)/q_i(x_i))
\end{align}
where $r(x_i)$ is acceptance probability of token $x_i$, $\mathrm{Random}(0,1)$ is a random number between 0 and 1. If $V_{\mathrm{SD}}(x_i)>0$, we accept the draft token and verify the next one, otherwise we reject it and sample a new token from the distribution:
\begin{equation}
\label{SD resample distribution}
\pi(x_{i}) = \frac{\max(0, p_i(x_{i}) - q_i(x_{i}))}{\sum_{y\in\mathcal{V}} \max(0, p_i(y) - q_i(y))}
\end{equation}
and discard the draft tokens $\{x_{i+1}, \ldots, x_{\gamma} \}$. The summation is over the whole vocabulary $\mathcal{V}$.

Speculative decoding has been proven to generate the same distribution as the target model theoretically~\cite{speculative_deepmind,speculative_google}.

\paragraph{Mentored Decoding.}
Alternative approaches \citep{faster_cascades, mentored_decoder} trade off the consistency of the distribution for a higher acceptance rate by adjusting the distance between the sampling and the target distribution flexibly. In Mentored Decoding~\cite{mentored_decoder}, the draft token $x_i$ is accepted if $V_{\mathrm{MD}}(x_i)>0$:
\begin{equation}
\label{lossy verify}
V_{\mathrm{MD}}(x_{i}) = \min(1, \frac{p_i(x_{i})}{\alpha q_i(x_{i})}) - \mathrm{Rand}(0,1)
\end{equation}
otherwise, we reject $x_i$ and sample a new token from distribution $\pi(x_i)$
\begin{equation}
\label{lossy resample}
\pi(x_{i}) = \frac{\max(0, p_i(x_{i})/\beta - q_i(x_{i}))}{1-\sum_{y\in\mathcal{V}}{p_i(y) \min(1, \frac{p_i(x_{i})}{\alpha q_i(x_{i})})}}
\end{equation}
where the coefficients of $\alpha$ and $\beta$ are calculated by iteratively optimizing to make the KL divergence between $q_i(x)$ and $p_i(x)$ fall within the interval $[D - \gamma, D + \gamma]$ \cite{mentored_decoder}, and $D$ and $\gamma$ are predefined thresholds of the distribution distance and tolerance.

\subsection{Consultant Decoding}
The perspective of CD is to validate the accuracy of the draft tokens using NLL of the target model, rather than to assess them based on the likelihood ratio of the target and the draft modls. 
We can formalize the verification method as a token-wise prediction NLL, to align with training loss as illustrated in the \Cref{fig:problem_SD}. 
As shown in Equation~\ref{consult decoding verify}, we treat the draft tokens as pseudo ground truth and compute the negative log-likelihood with the prediction probability of the target model, where $\varepsilon$ is the threshold of the accuracy of the draft tokens. The underlying rationale of this method is that LLMs (target models) learn the vocabulary distribution for a given context from the training data, thereby utilizing their outstanding performance to assess the probability of predicted tokens in drafts.

\begin{equation}
\label{consult decoding verify}
V_{\textrm{CD}}(x_{i}) = \varepsilon - (-\log(p_i(x_{i})))
\end{equation}

If $V_{\textrm{CD}}(x_i) > 0$, the draft token $x_{i}$ is accepted, otherwise we sample a new token from the distribution given by the target model. This approach eliminates the need for additional adjustments or modifications. By relying on the LLMs distribution, we preserve the integrity of its predictions and maintain coherence in the generated output. 

\begin{algorithm}[t]
\caption{Consultant Decoding}
\label{alg:consultant-decoding}
\begin{algorithmic}[1]
\REQUIRE{Small draft model $Q$, large target model $P$, threshold $\varepsilon$, EMA decay weight $\beta$, prefix tokens $x_{\leq t}$.} 
\FOR{$i = 1$ \textbf{to} $\gamma$}
\STATE $q_i(\cdot) \gets Q(x_{\leq t}, x_1, \dots, x_{i-1})$
\STATE $x_i, q_i(x_i) \sim q_i(\cdot) $
\ENDFOR
\STATE $p_1(\cdot), \dots, p_{\gamma+1}(\cdot) \gets P(x_{\leq t}, x_1, \dots, x_{\gamma})$
\STATE Set $n = 0$.
\FOR{$i = 1$ \textbf{to} $\gamma$}
\STATE $p_i(x_i) \gets p_i(\cdot), x_i$
\STATE $r_i=\mathrm{EMA}_{\beta}(r_{i-1}, -\log(p_i(x_i)))$
\IF{ $r_i \leq \varepsilon$ }
\STATE $n=i$
\ELSE
\STATE \textbf{break}
\ENDIF
\ENDFOR
\STATE $x'_{n+1} \gets p_{n+1}(\cdot)$
\STATE $r_{n+1} = \mathrm{EMA}_{\beta}(r_n, -\log(p_{n+1}(x'_{n+1})))$
\STATE \textbf{Return} $[x_{\leq t}, x_{1}, \dots, x_{n},x'_{n+1}], r_{n+1}$
\end{algorithmic}
\end{algorithm}

Furthermore, Perplexity (PPL) is a common measure used to estimate the likelihood of a given sentence based on the predicted tokens as studied in \cite{lu2023makes}, which can be formalized as $PPL = e^{\frac{1}{N}\sum(-\log(p_i))}$, where N is sentence length. It is obvious that the acceptance criteria in Consultant Decoding can be understood as a specific case where the sentence length $N = 1$. 

Given this interpretation, it is natural to introduce the smooth version of the Consultant Decoding. As shown in Equation~\ref{block consult decoding}, we utilize the exponential moving average (EMA) to smooth the token's NLL and take the context NLL into account. The parameter $\beta$ allows for a flexible trade-off between the ratios of context and the current NLL.

\begin{align}
\label{block consult decoding}
&V(x_{i}) =  \varepsilon - \mathrm{EMA}_{\beta}(r_i = -\log(p_i(x_{i}))) \\
&\mathrm{EMA}_{\beta}(r_i) = \beta\mathrm{EMA}_{\beta}(r_{i-1}) + (1-\beta)  r_i
\end{align}

Algorithm ~\ref{alg:consultant-decoding} provides a sketch of the Consultant Decoding implementation.

\subsection{How to Determine Generic Threshold $\varepsilon$}
\label{the generic threshold}
Except for SD, other verification methods have not been widely used, as it is challenging to determine thresholds that preserve inference accuracy across various LLMs and tasks.
As discussed in Section~\ref{sec:intro}, our CD provides a new perspective to address this issue. Specifically, the parameter $\varepsilon$ in CD is related to the convergence loss of target model during training. Therefore, we can set $\varepsilon$ to the convergence loss. For open-source models where convergence loss is not readily available, we estimate it using the Chinchilla scaling law proposed by \citet{hoffmann2022training}, as depicted in Equation~\ref{scalling law},
\begin{equation}
\label{scalling law}
L(N, D) = E + \frac{A}{N^{\alpha}} + \frac{B}{D^{\beta}}
\end{equation}
where $E = 1.69, A = 406.4, B = 410.7, \alpha = 0.34$, and $\beta = 0.28$. $N$ and $D$ are the model scale and the training data size of LLMs, respectively. Based on these calculations, we approximate the convergence loss as 2.0 for several models, and set $\varepsilon=2.0$  without additional parameter tuning. Although Chinchilla scaling law was originally fit on different LLMs, and thus $\varepsilon=2.0$ is a rough estimate, we find it performs reasonably well in most scenarios.

\subsection{Why Generic Threshold $\varepsilon$ can Preserve Accuracy of LLMs?}
\label{sec:why generic threshold can preserve accs}
\begin{figure}[t]
  \includegraphics[width=0.95\linewidth]{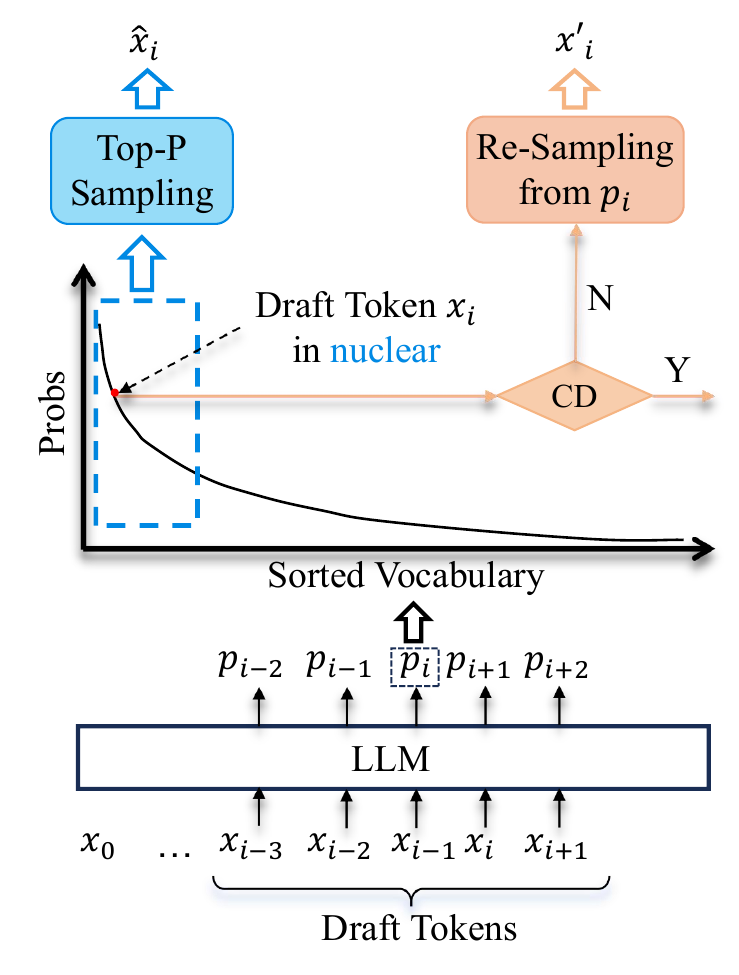}
  \caption{The comparison between Consultant Decoding and Top-P sampling. $x_i$, $x'_i$, and $\hat{x}_i$ denote draft token, resample token by CD, and sample token by Top-P, respectively.}
  \label{fig:CD_compare_topp}
\end{figure}

We can explain why the generic threshold of CD preserves target model accuracy through two key aspects: \textbf{1) Accuracy:} The CD threshold is aligned with the convergence loss of target model training, ensuring that predicted tokens reflect the distribution of the training dataset. This alignment allows CD to filter the correct tokens with high precision, thereby maintaining the overall performance of the target model. \textbf{2) Consistency:} The tokens accepted by CD are drawn from the set of results produced by top-p sampling of the target model. This indicates that when CD accepts a token, it corresponds to a likely output the target model would generate in the same context. Figure~\ref{fig:CD_compare_topp} visualizes the relationship between Top-P sampling and CD, showing how Top-P sampling achieves high-precision text prediction by sampling from a high-quality dynamic nucleus~\cite{holtzman2019ppl}.
As $\beta$ in Equation~\ref{block consult decoding} approaches 0, the verification formula for CD can be reformulated as $p(x_i)-e^{-\varepsilon}>0$. So, when high-quality dynamic nucleus is set by $1-e^{-\varepsilon}$, CD shares similar property of Top-P sampling, and its verification method is equivalent to determining whether a draft token lies within the high-quality dynamic nucleus of the target model, thus also enabling high-precision text prediction. Intuitively, the small draft model can be viewed as a sampler that sample from the high-quality dynamic nucleus of the large target model, thus combining the strengths of both target model and draft model. As shown in Figure~\ref{fig:acc_threshold}, appropriately expanding the size of the dynamic nucleus does not significantly degrade model performance on GSM8K.

\section{Experiments}
\label{experiments}
\paragraph{Models and Architectures.}
The architectures of speculative decoding can be broadly classified into two types: independent drafting and self-drafting~\cite{xia2024unlocking}. In independent drafting, a smaller, separate model is used as the draft model. In contrast, self-drafting utilizes the LLMs itself to efficiently generate draft tokens, making it a practical alternative when a dedicated draft model is not readily available.  In the independent drafting setting, we use Qwen2.5~\cite{yang2024qwen25} (a 72B target model, with 0.5B, 1.5B, 3B, and 7B draft models), and Llama3.1~\cite{dubey2024llama3} (a 70B target model, with an 8B draft model, shown in Appendix~\ref{sec:results_llama3.1}) models. In the self-drafting setting, experiments are conducted on EAGLE-2, which is one of the most effective methods for inference speedup currently available.

\paragraph{Tasks and Metrics.}
We evaluate the performance of verification methods across several tasks: mathematical reasoning GSM8K~\cite{cobbe2021gsm8k}, code generation HumanEval~\cite{chen2021humaneval}, multi-turn dialogue MT-Bench~\cite{zheng2023mtbench}, and instruction following AlpacaEval~\cite{li2023alpacaeval}. 
For each task, we adopt the corresponding evaluation metric used in the literature~\cite{yang2024qwen25} for performance evaluation.
The details of tasks and metrics are described in the following:

\begin{itemize}
\item \textbf{GSM8K}. GSM8K is designed to evaluate the mathematical reasoning abilities of LLMs, consisting of 1319 questions. We following \cite{wei2022chain-of-thought} using 8-shot, as shown in Table~\ref{tab:gsm8k-8-shot}, and step-by-step solution leading to the correct final answer. We use accuracy to evaluate model performance, which measures the correct rate of mathematical problem-solving.\\
\item \textbf{HumanEval}. HumanEval is used to evaluate the ability of models to generate python code based on natural language. Each sample includes a function signature (with parameter names and type hints) and a docstring (natural language specification of the function's intended behavior). Then we adopt pass@1 as the metric to assess the correctness of the model's single-attempt code generation.\\
\item \textbf{MT-Bench}. MT-Bench is a multi-turn dialogue benchmark, containing 80 samples. Responses are scored (1-10) by GPT-4o for helpfulness, relevance, accuracy, depth, creativity, and level of detail of the response, which is used by Medusa~\cite{cai2401medusa}. \\
\item \textbf{AlpacaEval}. AlpacaEval is a benchmark for evaluating the instruction-following capabilities of language models. It focuses on assessing how well models can generate helpful, high-quality responses based on user instructions. We follow \cite{li2023alpacaeval} and use the Win Rate as the evaluation metric, comparing model outputs against a baseline (i.e., text-davinci-003) via GPT-4o judgments.
\end{itemize}

\begin{table*}
  \centering
  \begin{adjustbox}{max width=0.9\textwidth}
  \begin{tabular}{lc|cc|cc|cc|cc}
  \hline
  \multirow{2}{*}{\textbf{Model}} & \multirow{2}{*}{\textbf{Verify}} & \multicolumn{2}{c|}{\textbf{GSM8K}} & \multicolumn{2}{c|}{\textbf{HumanEval}} & \multicolumn{2}{c|}{\textbf{MT-Bench}} & \multicolumn{2}{c}{\textbf{AlpacaEval}} \\
  & & Accuracy & Speedup & Pass@1 & Speedup & Score & Speedup & Win-Rate & Speedup \\
  \hline
  \multicolumn{2}{c|}{Qwen2.5-0.5B} & 47.7\% & 4.83$\times$ & 39.4\% & 4.56$\times$ & 43.6\% & 4.52$\times$ & 29.7\% & 4.6$\times$ \\
  \multicolumn{2}{c|}{Qwen2.5-72B} & 100.0\% & 1.0$\times$ & 100.0\% & 1.0$\times$ & 100.0\% & 1.0$\times$ & 100.0\% & 1.0$\times$ \\
  \hline
  \multirow{3}{*}{\begin{minipage}[t]{0.1\linewidth}\centering Qwen2.5\\0.5B/72B \end{minipage}}
  & SD & 99.9\% & 2.22$\times$ & 100.7\%  & 2.18$\times$ & 100.0\% & 1.57$\times$ & 99.8\% & 1.40$\times$ \\
  & MD & 96.8\% & 2.54$\times$ & 97.9\% & 2.41$\times$ & 95.2\% & 1.77$\times$ & 97.5\% & 1.68$\times$ \\
  & CD & 96.1\% & \textbf{3.09$\times$} & 97.2\% & \textbf{3.04$\times$} & 95.43\% & \textbf{2.06$\times$} & 97.2\% & \textbf{2.02$\times$} \\
  \hline
  \multicolumn{2}{c|}{Qwen2.5-7B} & 95.6\% & 3.75$\times$ & 96.5\% & 3.59$\times$ & 90.0\% & 4.44$\times$ & 95.9\% & 3.48$\times$ \\
  \hline
  \multirow{3}{*}{\begin{minipage}[t]{0.1\linewidth}\centering Qwen2.5\\7B/72B \end{minipage}}
  & SD & 99.9\% & 2.09$\times$ & 100.0\% & 2.12$\times$ & 100.0\% & 1.68$\times$ & 99.4\% & 1.57$\times$ \\
  & MD & 98.7\% & 2.82$\times$ & 97.2\% & 2.91$\times$ & 95.8\% & 2.63$\times$ & 97.8\% & 2.39$\times$ \\
  & CD & 99.1\% & \textbf{3.00$\times$} & 100.7\% & \textbf{3.13$\times$} & 96.3\% & \textbf{2.83$\times$} & 98.1\% & \textbf{2.60$\times$} \\
  \hline
  \end{tabular}
  \end{adjustbox}
  \caption{Comparing verification methods for speedup ratio, where 0.5B, 7B, and 72B all represent model sizes (the smaller is the draft model and the larger is the target model). The Qwen2.5-72B achieves accuracy of 95.53 on GSM8K dataset, pass@1 metric of 86.59 on HumanEval dataset, and scores of 8.65 on MT-Bench datasets, and win-rate of 96.71 on AlpacaEval datasets, respectively. The absolute values of other performance are shown in Table~\ref{tab:results_for_ratio_ab}.}
  \label{tab:results_for_ratio}
\end{table*}

In our experiments, unless otherwise specified, we normalize the accuracy, pass@1, score, and win-rate metrics across all tasks relative to the target model's greedy decoding performance. This normalization facilitates a comparative analysis of metric variations.
To assess the inference speed, we employ the following metrics:

\begin{itemize}
\item \textbf{Walltime speedup ratio}: The speedup relative to standard autoregressive decoding of the target model.

\item \textbf{LLM-calls ratio}: The ratio of calls to the LLMs compared to autoregressive decoding. Reducing target model calls enables greater parallel tasks throughput.

\item \textbf{Average acceptance length}: The average ratio of tokens accepted per draft. 
\end{itemize}

\paragraph{Experiment Setting.}
We present two parameter configurations for verification methods:

\begin{itemize}
\item \textbf{Speed-Optimal}. An extensive search through numerous experiments is conducted to find the optimal speedup for each algorithm, with parameter settings detail in Appendix~\ref{sec:parameters-for-speed}. 

\item \textbf{Generic}.  The Generic setting is used for a fair comparison with SD, where no parameter adjustments are required. From our novel perspective, we set threshold $\varepsilon$ to 2.0 and the EMA decay $\beta$ to 0.2 for CD, and threshold $D$ to 0.3 with a temperature of 0.3 for MD\footnote{These are the default parameters provided by MD.}.
\end{itemize}

\subsection{Main Results}

\subsubsection{Walltime Speedup Ratio of CD}
To fully demonstrate the maximum acceleration ratio of different verification methods, we adjust parameters and draft lengths (see Appendix~\ref{sec:parameters-for-speed} for detalis), allowing a moderate performance decrease, within 5\% of the large target model, in prediction accuracy. As indicated in Table~\ref{tab:results_for_ratio}, the combination of Qwen2.5-0.5B/72B models achieves at least 2.02$\times$ speedup ratio in all datasets. Thanks to the fixed inference paths and code templates on mathematical reasoning and code generation datasets, CD achieves the highest inference acceleration of over 3$\times$, 
which is average faster by $~0.8\times$ and $0.59\times$ compared to SD and MD, respectively. On the more challenging MT-Bench and AlpacaEval datasets, CD achieves speedup ratios of $2.06\times$ and $2.02\times$ for the Qwen2.5-0.5B/72B model combination, and $2.83\times $and $2.60\times$ for the Qwen2.5-7B/72B model combination, respectively.

Note that due to the minimal performance difference between the Qwen2.5-7B and Qwen2.5-72B models on GSM8K and HumanEval datasets (<5\% of Qwen2.5-72B), the speedup ratios presented in Table~\ref{tab:results_for_ratio} approach the limit of these model combinations, i.e., accepting all draft tokens.  Although the experiment does not reveal the upper limit of CD, it suggests that with the advancement of LLMs and the enhancement of draft model prediction accuracy and speed, verification methods that accept more drafts are increasingly practical.

Furthermore, we provide the results of speedup ratio with Generic setting, in Appendix~\ref{sec:walltime speedup with generic setting}, to fairly compare with SD and support practical use.

\begin{table*}\small
  \centering
  \begin{adjustbox}{max width=0.9\textwidth}
  \begin{tabular}{ll|ccc|ccc|ccc}
  \hline
  \multirow{2}{*}{\textbf{Model}} & \multirow{2}{*}{\textbf{Verify}} & \multicolumn{3}{c|}{\textbf{GSM8K}} & \multicolumn{3}{c|}{\textbf{HumanEval}} & \multicolumn{3}{c}{\textbf{MT-Bench}} \\
  & & Accuracy & Speedup & LLM-calls & Pass@1 & Speedup & LLM-calls & Score & Speedup & LLM-calls \\
  \hline
  \multicolumn{2}{c|}{Qwen2.5-0.5B} & 47.7\% & 4.83$\times$ & 0.0\% & 39.4\% & 4.56$\times$ & 0.0\% & 43.6\% & 4.52$\times$ & 0.0\% \\
  \multicolumn{2}{c|}{Qwen2.5-72B} & 100.0\% & 1.0$\times$ & 100\% & 100.0\% & 1.0$\times$ & 100\% & 100.0\% & 1.0$\times$ & 100\% \\
  \hline
  \multirow{3}{*}{\begin{minipage}[t]{0.05\linewidth} Qwen2.5\\0.5B/72B \end{minipage}} & SD & 100.0\% & 1.69$\times$ & 11.7\% & 100.0\% & 1.59$\times$ & 11.8\% & 100.0\% & 0.80$\times$ & 22.0\% \\
  & MD & 99.1\% & 2.03$\times$ & 10.2\% & 99.3\% & 1.91$\times$ & 10.2\% & 97.9\% & 0.99$\times$ & 19.0\% \\
  & CD & 100.3\% & \textbf{2.39}$\times$ & \textbf{9.1\%} & 99.3\% & \textbf{2.23}$\times$ & \textbf{9.1\%} & 96.5\% & \textbf{1.25}$\times$ & \textbf{15.8\%} \\
  \hline
  \multicolumn{2}{c|}{Qwen2.5-7B} & 95.6\% & 3.75$\times$ & 0.0\% & 96.5\% & 3.59$\times$ & 0.0\% & 90.0\% & 4.44$\times$ & 0.0\% \\
  \hline
  \multirow{3}{*}{\begin{minipage}[t]{0.05\linewidth} Qwen2.5\\7B/72B \end{minipage}} & SD & 100.0\% & 1.78$\times$ & 9.0\% & 100.0\% & 1.74$\times$ & 8.5\% & 100.0\% & 1.01$\times$ & 14.1\% \\
  & MD & 100.5\% & 2.24$\times$ & 7.5\% & 99.3\% & 2.20$\times$ & 7.1\% & 98.6\% & 1.40$\times$ & 9.2\% \\
  & CD & 100.2\% & \textbf{2.53}$\times$ & \textbf{6.5\%} & 103.5\% & \textbf{2.49}$\times$ & \textbf{6.2\%} & 99.8\% & \textbf{1.75}$\times$ & \textbf{8.6\%} \\
  \hline
  \end{tabular}
  \end{adjustbox}
  \caption{Comparing verification methods with draft length setting to 20, 0.5B, 7B, and 72B all represent model sizes, where the smaller is the draft model and the larger is the target model. The absolute values of performance are shown in Table~\ref{tab:accents_ab}.}
  \label{tab:results_for_acclen}
\end{table*}

\subsubsection{LLM-calls}
Based on the Generic setting, we set the draft length to 20 to reduce the LLM-calls ratio (i.e., target models). 
Table~\ref{tab:results_for_acclen} illustrates that, across most tasks, the CD method achieves a lower LLM-calls ratio compared to SD and MD, while preserving prediction accuracy of the target model. For example, on GSM8K and HumanEval datasets, using the Qwen2.5-0.5B as the draft model and the Qwen2.5-72B as the target model, CD achieves the lowest LLM-calls ratio of 9.1\%.
On HumanEval dataset, with the combination of Qwen2.5-7B/72B model, using only LLM-calls ratio of 6.2\%, CD surpasses the prediction accuracy of target model by an additional 3.5\%. This means that compared to autoregressive decoding, the computation time of the LLMs is reduced to 6\%.
On MT-Bench dataset, which is a more challenging dataset, CD significantly outperforms SD and MD in the LLM-calls ratio, showing a reduction of 6.2\% compared to SD and 3.2\% compared to MD. 
This indicates that in real dialogue scenarios, CD is better at identifying the correctness of draft tokens.
We further validate this advantage of CD in a real-world business benchmark designed for practical industry applications. It can be regarded as a combination of multiple aspects not limited to reasoning, coding, multi-turn conversion and instruction following, as detailed in Appendix~\ref{sec:results on business dataset}.

\begin{figure}[t]
  \includegraphics[width=\columnwidth]{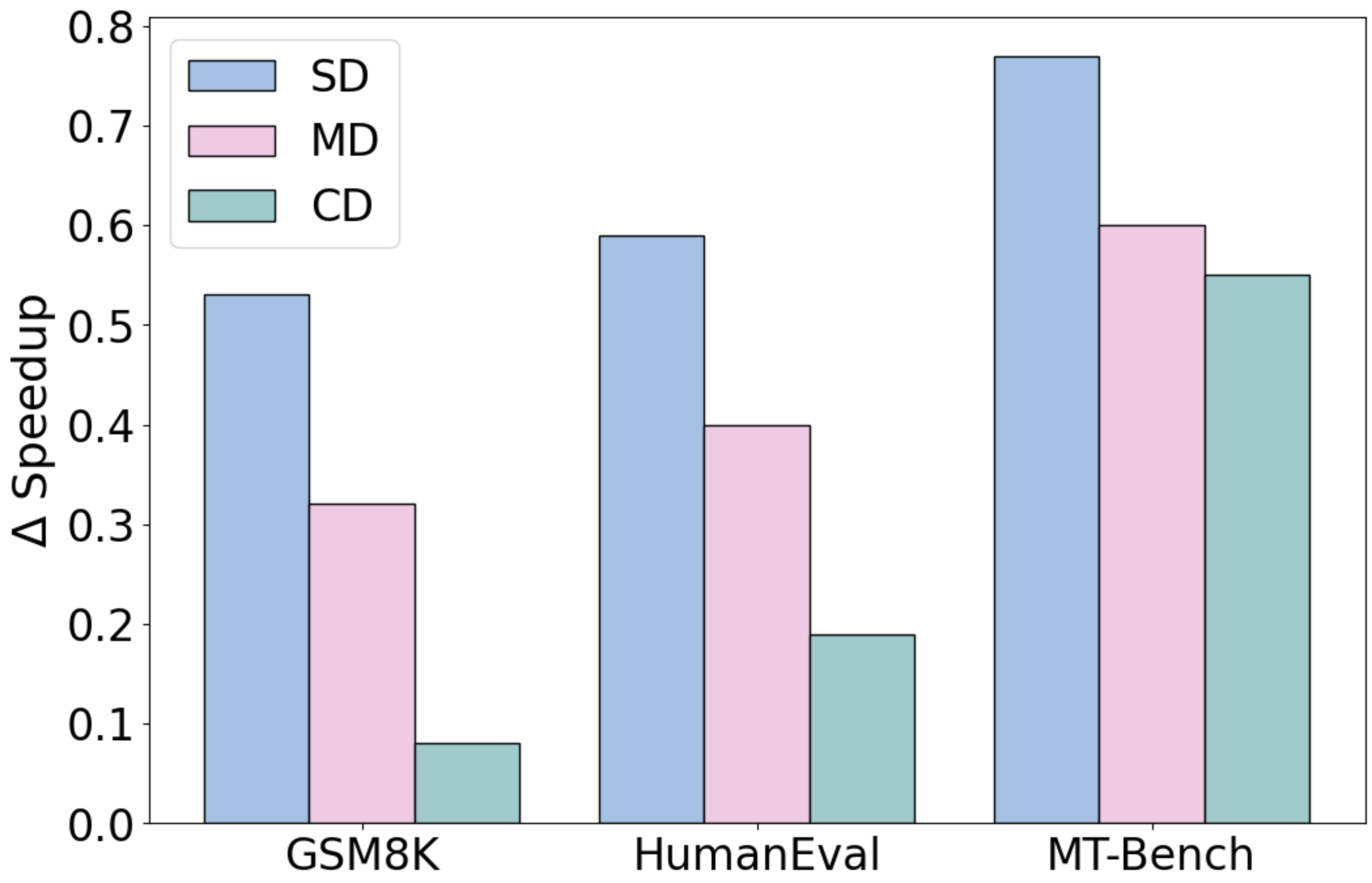}
  \caption{A comparison of the decrease in speedup ratios among SD, MD, and CD verification methods as the draft length increases from 6 to 20, using greedy decoding. Here target and draft models are 72B and 0.5B models, respectively.}
  \label{fig:delta_speedup} 
\end{figure}

\begin{table}
  \small
  \centering
  \begin{adjustbox}{max width=0.45\textwidth}
  \begin{tabular}{l|cc|cc}
  \hline
  \multirow{2}{*}{\textbf{Method}} & \multicolumn{2}{c|}{\textbf{HumanEval}} & \multicolumn{2}{c}{\textbf{MT-Bench}} \\
  & Pass@1 & Speedup & Score & Speedup  \\
  \hline
  Llama-3-70B & 100.0\% & 1.0$\times$ & 100.0\% & 1.0$\times$ \\
  Eagle-2-70B & 100.0\% & 3.43$\times$ & 100.0\% & 3.18$\times$ \\
  Eagle-2-70B-CD & 100.8\% & \textbf{3.66}$\times$ & 98.2\% & \textbf{3.51}$\times$ \\
  \hline
  \end{tabular}
  \end{adjustbox}
  \caption{Generalization of CD on EAGLE-2 structure. The Llama-3-70B achieve pass@1 accuracy of 76.83 on HumanEval, a score of 7.4 on MT-Bench.}
  \label{tab:results_for_structure}
\end{table}

\begin{table}
  \small
  \centering
  \begin{adjustbox}{max width=0.45\textwidth}
  \begin{tabular}{l|cc|cc}
  \hline
  \multirow{2}{*}{\textbf{Method}} & \multicolumn{2}{c|}{\textbf{HumanEval}} & \multicolumn{2}{c}{\textbf{MT-Bench}} \\
  & Pass@1 & Speedup & Score & Speedup  \\
  \hline
  Baseline & 100.0\% & 1.0$\times$ & 100.0\% & 1.0$\times$ \\
  + CD & 103.52\% & \textbf{2.51$\times$} & 99.8\% & \textbf{1.75$\times$} \\
  + CD (w/o Smooth) & 103.52\% & 2.40$\times$ & 99.3\% & 1.70$\times$ \\
  \hline
  \end{tabular}
  \end{adjustbox}
  \caption{Ablation study of Consultant Decoding with Qwen2.5-7B/72B.}
  \label{tab:results_for_smooth}
\end{table}

\subsubsection{CD's Resilience to Draft Length}
Essentially, the draft length impacts the overall speedup of synergistic decoding in a complex way, thus tweaking draft length to achieve reasonable balance of performance and inference speedup is 
notoriously challenging. We study how CD method is influenced by draft length via fixing the collaboration performance to around $100\%$.
As shown in ~\Cref{fig:delta_speedup}, when the draft length increases from 6 to 20, the speedup ratios of SD and MD on GSM8K and HumanEval decrease by 0.53$\times$, 0.59$\times$, and 0.32$\times$, 0.4$\times$, respectively. This is mainly because a larger proportion of draft tokens may be inappropriately rejected, which consumes a significant amount of time.
In contrast, our CD, which benefits from a more accurate assessment of draft token correctness, experienced only a slight reduction in the speedup ratio by 0.08× and 0.19×, respectively.

For the more challenging MT-Bench dataset, as illustrated in ~\Cref{fig:delta_speedup} and ~\Cref{tab:results_for_acclen}, CD shows the smallest decrease in speedup ratio while still achieving the highest speedup of 1.25$\times$. In contrast, SD performs worse than the standard autoregressive decoding used in LLMs.
Thus, our CD approach is more resilient to draft length variations in terms of maintaining the speedup ratio.

\subsubsection{Results on Self-drafting Decoding}
To demonstrate the generalization of CD, we replace the verification method in EAGLE-2 and compare the speedup ratios and accuracy. As presented in Table ~\ref{tab:results_for_structure}, CD shows strong generalization across benchmarks. Specifically, when applied to HumanEval and MT-Bench, EAGLE-2 with CD verification, the speed is improved by 0.23$\times$ and 0.33$\times$, respectively, with nearly no performance degradation. Additional analysis and discussion are provided in Appendix~\ref{sec:discussion_eagle}.

\begin{table*}[t!]\centering
\begin{minipage}{1.0\columnwidth}\vspace{0mm} \centering
\begin{tcolorbox} [grow to left by=4.8cm, grow to right by=4.8cm]
    \centering
   
      \footnotesize
    \begin{tabular}{l p{0.8\columnwidth}}
\textbf{Question:} & A food caterer was told to prepare gourmet hot dogs for 36 guests. While most people would only eat one hotdog, he prepared enough for half of the guests to be able to have two hotdogs. However, 40 guests showed up, and everyone wanted a second hotdog. How many guests did not get a second hotdog? \\
     & \hrulefill \\
\textbf{SD} & First, (\textcolor{gray}{ we}) \textcolor{blue}{ let}'s determine how many hotdogs (\textcolor{gray}{ were}) \textcolor{blue}{ the} caterer prepared (\textcolor{gray}{ for}) \textcolor{blue}{.} (\textcolor{gray}{ Since}) \textcolor{blue}{ The} caterer (\textcolor{gray}{ prepared}) \textcolor{blue}{ was} told to prepare (\textcolor{gray}{ gourmet}) \textcolor{blue}{ enough} for (\textcolor{gray}{ half}) \textcolor{blue}{ }36 guests, (\textcolor{gray}{ and}) \textcolor{blue}{ with} half of them (\textcolor{gray}{ being}) \textcolor{blue}{ able} to have two hotdogs. (\textcolor{gray}{ So}) \textcolor{blue}{ This} means (\textcolor{gray}{ }) \textcolor{blue}{:} (\textcolor{gray}{[}) \textcolor{blue}{-} Half of (\textcolor{gray}{ }) \textcolor{blue}{ the} guests (\textcolor{gray}{ =}) \textcolor{blue}{ (}18 guests) would have (\textcolor{gray}{ two}) \textcolor{blue}{ }2 hotdogs each (\textcolor{gray}{,}) \textcolor{blue}{:} (\textcolor{gray}{ }) \textcolor{blue}{ (} (\textcolor{gray}{1}) \textcolor{blue}{ }18 $\times$ 2 = 36 ) hotdogs. - The other half (\textcolor{gray}{ of}) \textcolor{blue}{ (}18 guests) would have 1 hotdog each: \( 18 \times 1 = 18 \) hotdogs.

So, the total number of hotdogs prepared is (\textcolor{gray}{ (}) \textcolor{blue}{:} $36$ (\textcolor{gray}{ $\backslash$}) \textcolor{blue}{ +} 18 = 54

 (\textcolor{gray}{Next}) \textcolor{blue}{Now}, 40 guests showed up, and (\textcolor{gray}{ everyone}) \textcolor{blue}{ each} (\textcolor{gray}{ of}) \textcolor{blue}{ wanted} a second hotdog. This means (\textcolor{gray}{ }) \textcolor{blue}{ each} (\textcolor{gray}{ of}) \textcolor{blue}{ guest} (\textcolor{gray}{ who}) \textcolor{blue}{ wanted} 2 hotdogs (\textcolor{gray}{,}) \textcolor{blue}{.} (\textcolor{gray}{ So}) \textcolor{blue}{ The} total number of hotdogs (\textcolor{gray}{ served}) \textcolor{blue}{ needed} for 40 guests is: $ 40 \times 2 = 80 $

 (\textcolor{gray}{Since}) \textcolor{blue}{The} (\textcolor{gray}{ number}) \textcolor{blue}{ cater}er (\textcolor{gray}{ prepared}) \textcolor{blue}{ only} (\textcolor{gray}{ prepared}) \textcolor{blue}{ had} 54 hotdogs, so the number of hotdogs (\textcolor{gray}{ not}) \textcolor{blue}{ short} is:$80 - 54 = 26$

 \underline{(\textcolor{gray}{Therefore}) \textcolor{blue}{Since} (\textcolor{gray}{ }) \textcolor{blue}{ each} guest (\textcolor{green}{ who}) \textcolor{blue}{ needs} 2 hotdogs (\textcolor{gray}{,}) \textcolor{blue}{ to} (\textcolor{gray}{ have}) \textcolor{blue}{ be} (\textcolor{gray}{ able}) \textcolor{blue}{ satisfied}, the number of} \underline{guests who did not get a second hotdog is: $\frac{26}{2} = 13$}

 (\textcolor{gray}{So}) \textcolor{blue}{The} answer is 13 (\textcolor{gray}{ guests}) \textcolor{blue}{.} \#\#\#\# 13 \\
     & \hrulefill \\
\textbf{CD} & First, we need to determine how many hotdogs (\textcolor{gray}{ were}) \textcolor{blue}{ the} caterer prepared. (\textcolor{gray}{ Since}) \textcolor{blue}{ The} caterer prepared enough for half of the (\textcolor{gray}{ guests}) \textcolor{blue}{ }36 guests (\textcolor{gray}{,}) \textcolor{blue}{ to} have two hotdogs each. So, half of 36 guests is 36 / 2 = 18 guests. (\textcolor{gray}{ This}) \textcolor{blue}{ These} 18 guests (\textcolor{gray}{ each}) \textcolor{blue}{ would} have 2 hotdogs each, so (\textcolor{gray}{ they}) \textcolor{blue}{ the} (\textcolor{gray}{ total}) \textcolor{blue}{ cater}er prepared 18 * 2 = 36 hotdogs (\textcolor{gray}{.}) \textcolor{blue}{ for} (\textcolor{gray}{ them}) \textcolor{blue}{ these} guests.

 (\textcolor{gray}{Next}) \textcolor{blue}{The} (\textcolor{gray}{ cater}) \textcolor{blue}{ remaining} (\textcolor{gray}{ guests}) \textcolor{blue}{ }18 guests (\textcolor{gray}{,}) \textcolor{blue}{ would} have  (\textcolor{red}{2}) \textcolor{blue}{1} hotdog each, so the caterer prepared 18 * 1 = 18 hotdogs for these guests.

 (\textcolor{gray}{Now}) \textcolor{blue}{In} total, the caterer prepared 36 + 18 = 54 hotdogs.

Now, 40 guests showed up, and (\textcolor{gray}{ everyone}) \textcolor{blue}{ each} (\textcolor{gray}{ of}) \textcolor{blue}{ wanted} a second hotdog. (\textcolor{gray}{ So}) \textcolor{blue}{ This} means 40 * 2 = 80 hotdogs were (\textcolor{gray}{ consumed}) \textcolor{blue}{ needed}.

The number of hotdogs that were not (\textcolor{gray}{ prepared}) \textcolor{blue}{ available} is (\textcolor{gray}{ the}) \textcolor{blue}{ } (\textcolor{red}{5}) \textcolor{blue}{8}0 - 54 = 26 hotdogs.

 \underline{(\textcolor{gray}{Therefore}) \textcolor{blue}{Since} each guest who did not get a second hotdog (\textcolor{gray}{ would}) \textcolor{blue}{ is} (\textcolor{gray}{ counted}) \textcolor{blue}{ missing} one hotdog,} \underline{the number of guests who did not get a second hotdog is 26.}

The answer is 26. \#\#\#\# 26 \\
    \end{tabular}
\end{tcolorbox}
\end{minipage}
\caption{The example of comparing SD and CD with Qwen2.5-0.5B/7B, on GSM8K. The \textcolor{blue}{blue text} represents the text modified by target model, \textcolor{gray}{gray text} denotes candidate text generate by draft model, \textcolor{red}{red text} and \textcolor{green}{green text} are critical modifications and incorrect modifications, respectively. \underline{The underline text} is different thought chain between SD and CD.}
\label{tab:case-study}
\end{table*}

\subsection{Ablation Study}
\subsubsection{Threshold $\varepsilon$}
\label{sec:threshold}

\begin{figure}[t]
  \includegraphics[width=1.0\linewidth]{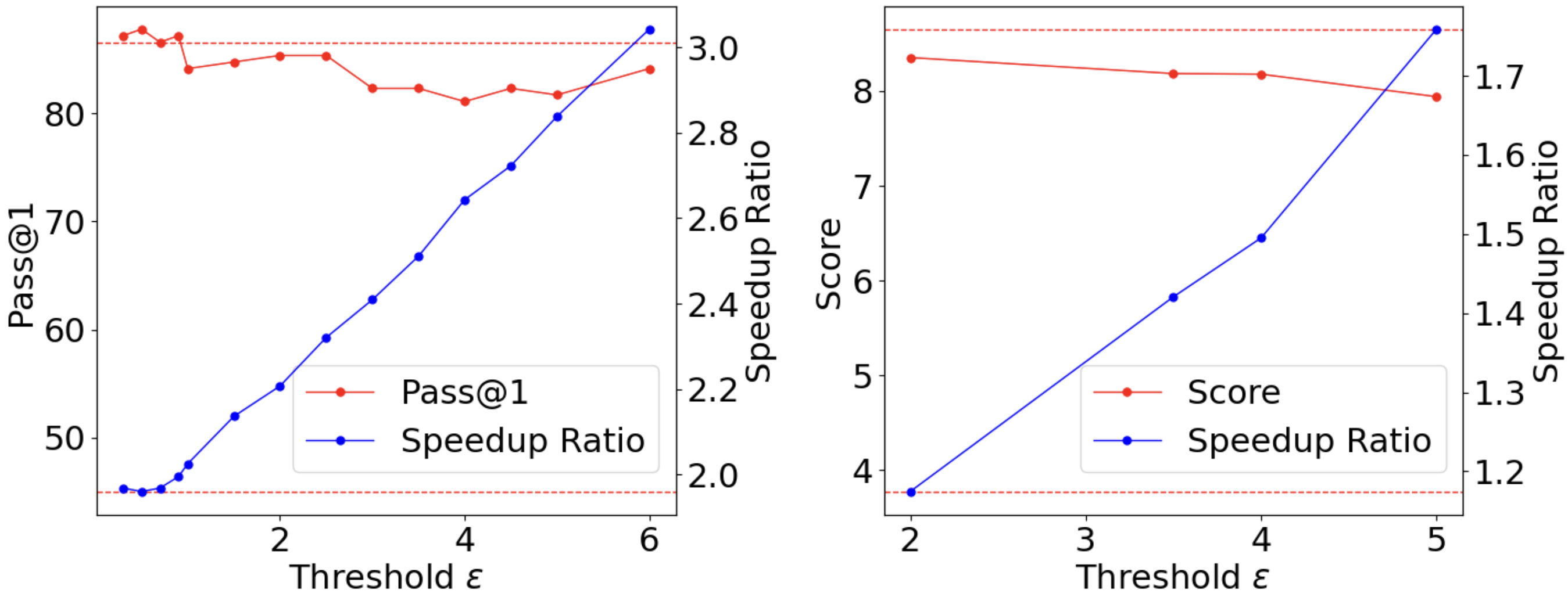}
  \caption{Ablation study of threshold $\varepsilon$ of CD with Qwen2.5-0.5B/72B, on HumanEval and MT-Bench. The red dashed lines are the inference accuracy of the target and draft models.}
  \label{fig:acc_threshold}
\end{figure}

We evaluate the performance of CD by varying the $\varepsilon$ parameter. As depicted in Figure~\ref{fig:acc_threshold}, the performance of CD does not significantly deteriorate with a more relaxed threshold. Specifically, on GSM8K dataset, performance remains largely stable as $\varepsilon$ increases from 2 to 6.  On MT-Bench, an 8\% performance decrease is observed, yet it still outperforms smaller models. 
In practice, the optimal $\varepsilon$ can be selected based on the trade-off between performance and efficiency required.

\subsubsection{Smooth Verification}
Smooth verification is used to balance the contribution of the current prediction and context tokens. If the tokens in the context have a low NLL, as calculated by Equation~\ref{consult decoding verify}, we can somewhat relax the constraints on the current token. This helps reduce accidental errors in the model. Table~\ref{tab:results_for_smooth} presents comparative results that indicate the impact of using smooth verification. Smooth verification has been shown to increase the acceleration ratio by 0.11$\times$ on HumanEval and by 0.05$\times$ on MT-Bench, without compromising the model performance.

\subsection{Case Study}
Consultant Decoding evaluates the correctness of draft tokens using the target model, rather than strictly aligning the LLMs distribution. This approach combines the strengths of both the target and draft models for more accurate predictions.
As shown in Table~\ref{tab:case-study}, we compare SD and CD on GSM8K dataset. In SD, the target model incorrectly connects "the number of hotdogs short" and "each guest needs 2 hotdogs", leading to an erroneous prediction. In contrast, CD allows the target model to follow the reasoning path of the draft model, resulting in a correct prediction. This illustrates why CD can surpass the performance upper bound of speculative decoding, thereby improving the overall performance of the target model.

\subsection{Scalability}
\label{question2}
\begin{figure}[t]
  \includegraphics[width=\linewidth]{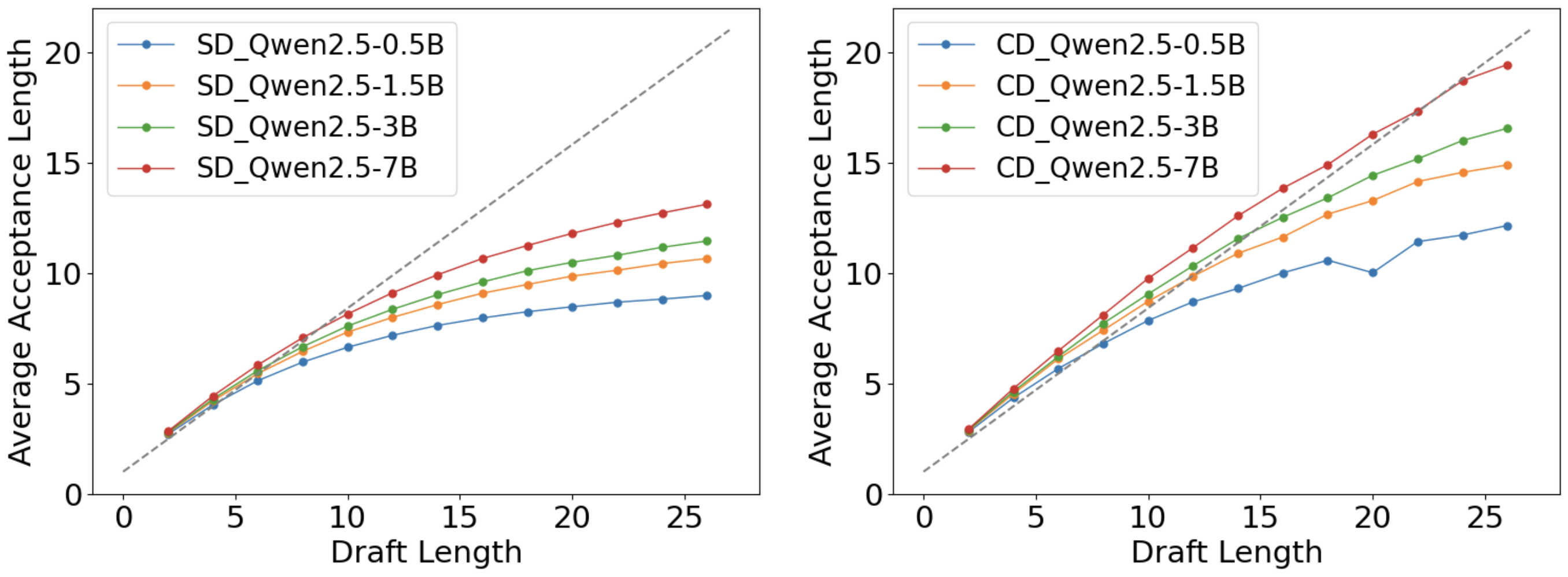}
  \caption{The figure shows the scalability of average acceptance length on HumanEval. The target model is Qwen2.5-72B, and draft models are Qwen2.5-0.5B, 1.5B, 3B, 7B. The gray dashed lines extend from the coordinates (0,1) to (27,24).}
  \label{fig:draft_length scaling}
\end{figure}

To investigate why CD can reduce the LLM-calls ratio without significantly compromising the speedup ratio, we fix the parameters of CD as in Generic setting to ensure high-quality predictions. We then examine the Average Acceptance Length (AAL) of draft tokens with various draft lengths.

The results, shown in Figure~\ref{fig:draft_length scaling}, reveal that as the draft length increases, the growth rates of AAL for SD and CD slow down. Notably, the AAL of SD saturates quickly, whereas the AAL of CD shows a sharper increase.
Furthermore, as the draft model size increases, the AAL of CD grows more significantly than that of SD, indicating a more pronounced scaling behavior. 
For instance, with a 7B draft model, the growth of AAL for SD from draft length 2 to 26 is 10.28, while for CD it is 16.51.
In other words, compared to SD, CD achieves higher AAL and a lower LLM-calls ratio more easily by increasing the draft model size.

\section{Conclusion}
This paper develops Consultant Decoding (CD), a novel decoding method based on NLL, that improves speculative decoding for model inference speedup. Unlike traditional methods that trade efficiency for performance, CD aligns with the convergence loss of target model training, achieving higher speedup without compromising or even enhancing the target model performance by theoretical analysis and case study. 
Comprehensive experiments show the acceleration advantage, generalization, resilience and scalability of CD. Furthermore, we provide a comparison of CD with Top-P sampling, providing a theoretical basis for how CD maintains the inference accuracy of the target model with a threshold $\varepsilon$ of 2.0. This advantage provides convenience for the use of CD in practical scenarios.
The generalization of CD to other architectures (e.g., EAGLE-2) further highlights its robustness and scalability.

\section*{Limitations}
Our propose perspective aligns the prediction NLL of the draft token with the convergence loss of target model training, facilitating the replacement of the strict SD verification methd with a fixed-threshold CD ($\varepsilon=2.0$), which achieves significant inference acceleration. 
However, this fixed threshold is not optimal. Figure~\ref{fig:acc_threshold} illustrates that increasing the threshold can further enhance the efficiency of CD without significantly compromising the performance of LLMs. The optimal threshold is highly dependent on both the questions and the combination of models, and determining this threshold during inference to fully leverage CD remains challenging. 
Moreover, another limitation is the occurrence of unnecessary modifications. While CD has reduced the cost of unnecessary modifications, as seen in Table~\ref{tab:case-study}, “Therefore” is changed to “Since”. Hence, improving the capability to distinguish the correct tokens is another direction for future study.

\bibliography{acl_latex}

\appendix

\section{Other Results}

\subsection{Results on Business Dataset}
\label{sec:results on business dataset}
As shown in Table~\ref{tab:results_for_business}, we present the performance of CD on the business dataset. To maintain high prediction accuracy, we employ the Generic setting with a draft length of 6. CD achieves the highest speedup ratio while preserving the prediction accuracy of the LLMs. For example, compared to SD, the speedup of CD is improved by 0.49$\times$ with only a 1.7\% decrease in score. Relative to MD, the speedup of CD is enhanced by 0.36$\times$, with a score increase of 2.8\%. In pursuit of a lower LLM-calls ratio, we utilize the Speed-Optimal setting with a draft length of 20. It is noteworthy that both MD and CD offer a trade-off between inference speed and prediction accuracy, however, at a similar speedup, CD exhibits a 16.1\% higher inference accuracy than MD. 

\begin{table}\small
  \centering
  \begin{adjustbox}{max width=0.9\textwidth}
  \begin{tabular}{l|l}
  \hline
  \textbf{Dataset} & \textbf{Prompt} \\
  \hline
  GSM8K & \{Instruction\}\textbackslash n Let's think step by step. At the \\ 
  & end, you MUST write the answer as an integer \\
  & after \'\#\#\#\#\'. \\
  \hline
  HumanEval & You are an intelligent programming assistant \\
  & to produce Python algorithmic solutions.\textbackslash n \\
  & Can you complete the following Python func- \\
  & -tion? \textbackslash n```python\textbackslash n \{Instruction\}\textbackslash n``` \\
  \hline
  MT-Bench & You are a helpful, respectful and honest assis- \\
  & -tant. Always answer as helpfully as possible, \\
  & while being safe. Your answers should not in- \\
  & -clude any harmful, unethical, racist, sexist, t- \\
  & -oxic, dangerous, or illegal content. Please en- \\
  & -sure that your responses are socially unbiased \\
  & and positive in nature. \textbackslash n\textbackslash n If a question does \\
  & not make any sense, or is not factually cohere- \\
  & -nt, explain why instead of answering someth- \\
  & -ing not correct. If you don't know the answer \\
  & to a question, please don't share false \\ 
  & information. \\
  \hline
  AlpacaEval & \{Instruction\} \\
  \hline
  \end{tabular}
  \end{adjustbox}
  \caption{The prompt corresponding to the dataset.}
  \label{tab:prompts}
\end{table}

\begin{table*}\small
  \begin{tcolorbox} [grow to left by=0cm, grow to right by=0cm]
  \textbf{Q:} There are 15 trees in the grove. Grove workers will plant trees in the grove today. After they are done, there will be 21 trees. How many trees did the grove workers plant today? Let's think step by step. At the end, you MUST write the answer as an integer after \'\#\#\#\#\'. \\
  \textbf{A:} We start with 15 trees. Later we have 21 trees. The difference must be the number of trees they planted. So, they must have planted 21 - 15 = 6 trees. The answer is 6.\textbackslash n \\
  
  \textbf{Q:} If there are 3 cars in the parking lot and 2 more cars arrive, how many cars are in the parking lot? Let's think step by step. At the end, you MUST write the answer as an integer after \'\#\#\#\#\'. \\
  \textbf{A:} There are 3 cars in the parking lot already. 2 more arrive. Now there are 3 + 2 = 5 cars. The answer is 5.\textbackslash n \\
  
  \textbf{Q:} Leah had 32 chocolates and her sister had 42. If they ate 35, how many pieces do they have left in total? Let's think step by step. At the end, you MUST write the answer as an integer after \'\#\#\#\#\'. \\
  \textbf{A:} Leah had 32 chocolates and Leah's sister had 42. That means there were originally 32 + 42 = 74 chocolates. 35 have been eaten. So in total they still have 74 - 35 = 39 chocolates. The answer is 39.\textbackslash n \\
  
  \textbf{Q:} Jason had 20 lollipops. He gave Denny some lollipops. Now Jason has 12 lollipops. How many lollipops did Jason give to Denny? Let's think step by step. At the end, you MUST write the answer as an integer after \'\#\#\#\#\'. \\
  \textbf{A:} Jason had 20 lollipops. Since he only has 12 now, he must have given the rest to Denny. The number of lollipops he has given to Denny must have been 20 - 12 = 8 lollipops. The answer is 8.\textbackslash n \\
  
  \textbf{Q:} Shawn has five toys. For Christmas, he got two toys each from his mom and dad. How many toys does he have now? Let's think step by step. At the end, you MUST write the answer as an integer after \'\#\#\#\#\'. \\
  \textbf{A:} He has 5 toys. He got 2 from mom, so after that he has 5 + 2 = 7 toys. Then he got 2 more from dad, so in total he has 7 + 2 = 9 toys. The answer is 9.\textbackslash n \\
  
  \textbf{Q:} There were nine computers in the server room. Five more computers were installed each day, from monday to thursday. How many computers are now in the server room? Let's think step by step. At the end, you MUST write the answer as an integer after \'\#\#\#\#\'. \\
  \textbf{A:} There are 4 days from monday to thursday. 5 computers were added each day. That means in total 4 * 5 = 20 computers were added. There were 9 computers in the beginning, so now there are 9 + 20 = 29 computers. The answer is 29.\textbackslash n \\
  
  \textbf{Q:} Michael had 58 golf balls. On tuesday, he lost 23 golf balls. On wednesday, he lost 2 more. How many golf balls did he have at the end of wednesday? Let's think step by step. At the end, you MUST write the answer as an integer after \'\#\#\#\#\'.\\
  \textbf{A:} Michael initially had 58 balls. He lost 23 on Tuesday, so after that he has 58 - 23 = 35 balls. On Wednesday he lost 2 more so now he has 35 - 2 = 33 balls. The answer is 33.\textbackslash n\\
  
  \textbf{Q:} Olivia has \$23. She bought five bagels for \$3 each. How much money does she have left? Let's think step by step. At the end, you MUST write the answer as an integer after \'\#\#\#\#\'.\\
  \textbf{A:} She bought 5 bagels for \$3 each. This means she spent 5 * \$3 = \$15 on the bagels. She had \$23 in beginning, so now she has \$23 - \$15 = \$8. The answer is 8.\textbackslash n\\
  \end{tcolorbox}
  \caption{The 8-shot prompt of GSM8K.}
  \label{tab:gsm8k-8-shot}
\end{table*}

\subsection{What is the maximum walltime speedup
ratio of CD on Generic setting?}
\label{sec:walltime speedup with generic setting}
To ensure a fair comparison with SD, where no parameter adjustment is required, and to support the practical application of CD, we teste the optimal speedup ratio for various varification methods based on the Generic setting with draft length setting to 6. As shown in Table~\ref{tab:results_for_ratio_preserving_acc}, CD achieves the best speedup ratio in all tasks while maintaining the prediction accuracy of LLMs. For example, with the combination of Qwen2.5-7B/72B, the speed up of CD exceeds SD by 0.41, 0.39, and 0.25, respectively, and exceeded MD by 0.26, 0.29, and 0.25, respectively, on GSM8K, HumanEval, and MT-Bench.

\begin{table}\small
  \centering
  \begin{adjustbox}{max width=0.9\textwidth}
  \begin{tabular}{lc|ccc}
  \hline
  \textbf{Setting} & \textbf{Verify} & Score & Speedup & LLM-calls  \\
  \hline
  \multirow{3}{*}{G-6} & SD & 100.0\% & 1.29$\times$ & 36.53\% \\
  & MD & 95.5\% & 1.42$\times$ & 30.42\% \\
  & CD & 98.3\% & \textbf{1.78$\times$} & 26.89\% \\
  \hline
  \multirow{3}{*}{S-20} & SD & 98.9\% & 0.79$\times$ & 28.45\% \\
  & MD & 67.4\% & 2.14$\times$ & 10.38\% \\
  & CD & 83.5\% & \textbf{2.15$\times$} & 10.57\% \\
  \hline
  \end{tabular}
  \end{adjustbox}
  \caption{Comparing verification methods on business dataset, where "G-6" and "S-20" denote the employment of the Generic setting with a draft length of 6 and Speed-Optimal setting with draft length of 20. }
  \label{tab:results_for_business}
\end{table}

\subsection{Results with Sampling Decoding}
Table~\ref{tab:results_for_sampling_decoding} presents a comparison of verification methods using sampling decoding for two combination of the Qwen2.5 model, Qwen2.5-0.5B/72B and Qwen2.5-7B/72B, based on a Generic setting with a draft length of 6 and temperature of 0.7. For both model sizes, the CD method exhibits a speedup of over 2$\times$ on GSM8K and HumanEval datasets, while maintaining accuracy above 99\%, even more than 100\%. 
This demonstrates that CD can reliably verify the accuracy of draft tokens, and as long as the draft model produces correct tokens, the decoding method may be selected based on specific application requirements.

\begin{table}\small
  \centering
  \begin{adjustbox}{max width=0.9\textwidth}
  \begin{tabular}{l|ccc}
  \hline
  \textbf{Datasets} & Accuracy & Speedup & LLM-calls  \\
  \hline
  \multicolumn{2}{c}{\small{\textit{Qwen2.5-0.5B/72B}}} & & \\
  \textbf{GSM8K} & 99.61\% & 2.25$\times$ & 18.78\% \\
  \textbf{HumanEval} & 101.41\% & 2.25$\times$ & 18.93\% \\
  \textbf{MT-Bench} & 97.54\% & 1.60$\times$ & 25.78\% \\
  \hline
  \multicolumn{2}{c}{\small{\textit{Qwen2.5-7B/72B}}} & & \\
  \textbf{GSM8K} & 100.16\% & 2.17$\times$ & 16.06\% \\
  \textbf{HumanEval} & 102.82\% & 2.16$\times$ & 16.01\% \\
  \textbf{MT-Bench} & 98.34\% & 1.82$\times$ & 18.59\% \\
  \hline
  \end{tabular}
  \end{adjustbox}
  \caption{Results of Consultant Decoding with sampling decoding, based on Generic setting with draft length of 6 and temperature of 0.7. }
  \label{tab:results_for_sampling_decoding}
\end{table}

\subsection{Results On Llama-3.1}
\label{sec:results_llama3.1}
We provide the results of Consultant Decoding with model combination of Llama-3.1-8B/70B, based on the Generic setting with greedy decoding. As shown in Table~\ref{tab:results_for_llama3.1}, the CD demonstrates excellent generalization on Llama3.1. For example, CD achieves the speedup ratio by 2.36$\times$, 2.38$\times$, and 2.22$\times$ on GSM8K, HumanEval and MT-Bench, respectively, while preserving the inference accuracy of target model. 
\begin{table}\small
  \centering
  \begin{adjustbox}{max width=0.9\textwidth}
  \begin{tabular}{l|ccc}
  \hline
  \textbf{Datasets} & Accuracy & Speedup & LLM-calls  \\
  \hline
  \textbf{GSM8K} & 99.05\% & 2.36$\times$ & 5.80\% \\
  \textbf{HumanEval} & 100.79\% & 2.38$\times$ & 5.45\% \\
  \textbf{MT-Bench} & 100.0\% & 2.22$\times$ & 7.48\% \\
  \hline
  \end{tabular}
  \end{adjustbox}
  \caption{Results of Consultant Decoding with combination of Llama-3.1-8B/70B, based on Generic setting, with draft length of 20 on GSM8K and HumanEval, draft length of 16 on MT-Bench.}
  \label{tab:results_for_llama3.1}
\end{table}

\section{Performance Analysis}
\subsection{Outperformance Analysis}
As shown in Table 2, the CD approach demonstrates better performance compared to the target model in three specific scenarios: (1) GSM8K with Qwen2.5-0.5B/72B, (2) GSM8K with Qwen2.5-7B/72B, and (3) HumanEval with Qwen2.5-7B/72B. Notably, the HumanEval case exhibits a notable improvement of 3.5\%.

We propose that when the target model encounters uncertainty (i.e., when multiple solutions with comparable probabilities exist), the SLM can effectively guide the decoding process toward higher-probability solutions based on its own modeling. As CD only accepts the draft model solution if it is high enough according to the target model, CD acts as a voting mechanism, allowing CD to combine the strength of the draft and the target model. This intriguing phenomenon presents a promising direction for future research.
\subsection{Error Analysis}
We examine 22 error samples from HumanEval and 56 from GSM8k, categorizing them into three distinct types:
\begin{itemize}
    \item \textbf{Target Model Errors:} Cases where the target model produce incorrect predictions.
    \item \textbf{Draft Model Errors:} Cases where the target model was correct but the draft model made mistakes.
    \item \textbf{CD Errors:} Cases where both models were correct individually, yet errors emerge from the CD process.
\end{itemize}

\begin{table}\small
  \centering
  \begin{adjustbox}{max width=0.9\textwidth}
  \begin{tabular}{l|ccc}
  \hline
  \textbf{Error Type} & HumanEval & GSM8K \\
  \hline
  \textbf{Target model errors} & 15 & 44 \\
  \textbf{Draft model errors} & 4 & 10 \\
  \textbf{CD errors} & 3 & 2 \\
  \textbf{Total errors} & 22 & 56 \\
  \hline
  \end{tabular}
  \end{adjustbox}
  \caption{The number of error samples of each type on HumanEval and GSM8K datasets.}
  \label{tab:error_number_cd}
\end{table}

As shown in Table~\ref{tab:error_number_cd}, our analysis reveals that:
\begin{itemize}
\item The predominant source of CD prediction errors originates from target model errors.
\item A minor proportion stems from either draft model errors or CD-specific errors.
\end{itemize}

We draw two key observations from these findings:
\begin{itemize}
\item Draft model errors that lead to CD failures indicate the existence of error tokens that challenge the LLMs identification capability.
\item CD errors likely arise from the language models' sensitivity to contextual variations generated through the draft-target model cooperation.
\end{itemize}
These insights suggest that future improvements in CD threshold optimization and LLMs robustness could effectively mitigate such error cases.

\begin{table*}
  \centering
  \begin{adjustbox}{max width=0.9\textwidth}
  \begin{tabular}{lc|cc|cc|cc|cc}
  \hline
  \multirow{2}{*}{\textbf{Model}} & \multirow{2}{*}{\textbf{Verify}} & \multicolumn{2}{c|}{\textbf{GSM8K}} & \multicolumn{2}{c|}{\textbf{HumanEval}} & \multicolumn{2}{c|}{\textbf{MT-Bench}} & \multicolumn{2}{c}{\textbf{AlpacaEval}} \\
  & & Accuracy & Speedup & Pass@1 & Speedup & Score & Speedup & Win-Rate & Speedup \\
  \hline
  \multicolumn{2}{c|}{Qwen2.5-0.5B} & 45.56 & 4.83$\times$ & 34.15 & 4.56$\times$ & 3.77 & 4.52$\times$ & 28.70 & 4.6$\times$ \\
  \multicolumn{2}{c|}{Qwen2.5-72B} & 95.53 & 1.0$\times$ & 86.59 & 1.0$\times$ & 8.65 & 1.0$\times$ & 96.71 & 1.0$\times$ \\
  \hline
  \multirow{3}{*}{\begin{minipage}[t]{0.1\linewidth}\centering Qwen2.5\\0.5B/72B \end{minipage}}
  & SD & 95.45 & 2.22$\times$ & 87.20 & 2.18$\times$ & 8.65 & 1.57$\times$ & 96.51 & 1.40$\times$ \\
  & MD & 92.50 & 2.54$\times$ & 84.76 & 2.41$\times$ & 8.23 & 1.77$\times$ & 94.28 & 1.68$\times$ \\
  & CD & 91.81 & \textbf{3.09$\times$} & 84.15 & \textbf{3.04$\times$} & 8.26 & \textbf{2.06$\times$} & 93.98 & \textbf{2.02$\times$} \\
  \hline
  \multicolumn{2}{c|}{Qwen2.5-7B} & 91.36 & 3.75$\times$ & 83.54 & 3.59$\times$ & 7.79 & 4.44$\times$ & 92.79 & 3.48$\times$ \\
  \hline
  \multirow{3}{*}{\begin{minipage}[t]{0.1\linewidth}\centering Qwen2.5\\7B/72B \end{minipage}}
  & SD & 95.45 & 2.09$\times$ & 86.59 & 2.12$\times$ & 8.65 & 1.68$\times$ & 96.13 & 1.57$\times$ \\
  & MD & 94.31 & 2.82$\times$ & 84.15 & 2.91$\times$ & 8.29 & 2.63$\times$ & 94.58 & 2.39$\times$ \\
  & CD & 94.64 & \textbf{3.00$\times$} & 87.20 & \textbf{3.13$\times$} & 8.33 & \textbf{2.83$\times$} & 94.84 & \textbf{2.60$\times$} \\
  \hline
  \end{tabular}
  \end{adjustbox}
  \caption{Comparing verification methods for speedup ratio, where 0.5B, 7B, and 72B all represent model sizes, where the smaller is the draft model and the larger is the target model. The Qwen2.5-72B achieve accuracy of 95.53 on GSM8K dataset, pass@1 metric of 86.59 on HumanEval dataset, and scores of 8.65 and 96.71 on MT-Bench and AlpacaEval datasets, respectively.}
  \label{tab:results_for_ratio_ab}
\end{table*}

\begin{table*}\small
  \centering
  \begin{tabular}{lc|ccc|ccc|ccc}
  \hline
  \multirow{2}{*}{\textbf{Model}} & \multirow{2}{*}{\textbf{Method}} & \multicolumn{3}{c|}{\textbf{GSM8K}} & \multicolumn{3}{c|}{\textbf{HumanEval}} & \multicolumn{3}{c}{\textbf{MT-Bench}} \\
  & & Accuracy & Speedup & $\tau$ & Pass@1 & Speedup & $\tau$ & Score & Speedup & $\tau$ \\
  \hline
  \multicolumn{2}{c|}{Qwen2.5-0.5B} & 45.56 & 4.83$\times$ & - & 34.15 & 4.56$\times$ & - & 3.77 & 4.52$\times$ & - \\
  \multicolumn{2}{c|}{Qwen2.5-72B} & 95.53 & 1.0$\times$ & - & 86.59 & 1.0$\times$ & - & 8.65 & 1.0$\times$ & - \\
  \hline
  \multirow{3}{*}{\begin{minipage}[t]{0.1\linewidth}\centering Q2.5\\-0.5B-72B \end{minipage}} 
  & SD & 95.53 & 1.69$\times$ & 8.55 & 86.59 & 1.59$\times$ & 8.50 & 8.65 & 0.80$\times$ & 4.54 \\
  & MD & 94.63 & 2.03$\times$ & 9.77 & 85.99 & 1.91$\times$ & 9.83 & 8.47 & 0.99$\times$ & 5.28 \\
  & CD & 95.83 & 2.39$\times$ & 10.99 & 85.97 & 2.23$\times$ & 10.97 & 8.35 & 1.25$\times$ & 6.34  \\
  \hline
  \multicolumn{2}{c|}{Qwen2.5-7B} & 91.36 & 3.75$\times$ & - & 83.54 & 3.59$\times$ & - & 7.79 & 4.44$\times$ & - \\
  \hline
  \multirow{3}{*}{\begin{minipage}[t]{0.1\linewidth}\centering Q2.5\\-7B-72B \end{minipage}}
  & SD & 95.53 & 1.78$\times$ & 11.18 & 86.59 & 1.74$\times$ & 11.79 & 8.65 & 1.01$\times$ & 7.08 \\
  & MD & 95.98 & 2.24$\times$ & 13.31 & 85.99 & 2.20$\times$ & 14.16 & 8.53 & 1.40$\times$ & 9.18 \\
  & CD & 95.68 & 2.53$\times$ & 15.37 & 89.63 & 2.49$\times$ & 16.25 & 8.63 & 1.75$\times$ & 11.66 \\
  \hline
  \end{tabular}
  \caption{Comparing verification methods with draft length setting to 20, 0.5B, 7B, and 72B all represent model sizes, where the smaller is the draft model and the larger is the target model. $\tau$ is the average acceptance length. }
  \label{tab:accents_ab}
\end{table*}

\begin{table*}\small
  \centering
  \begin{adjustbox}{max width=0.9\textwidth}
  \begin{tabular}{lc|cc|cc|cc}
  \hline
  \multirow{2}{*}{\textbf{Model}} & \multirow{2}{*}{\textbf{Verify}} & \multicolumn{2}{c|}{\textbf{GSM8K}} & \multicolumn{2}{c|}{\textbf{HumanEval}} & \multicolumn{2}{c}{\textbf{MT-Bench}} \\
  & & Accuracy & Speedup & Pass@1 & Speedup & Score & Speedup \\
  \hline
  \multicolumn{2}{c|}{Qwen2.5-0.5B} & 45.56 (47.7\%) & 4.83$\times$ & 34.15
 (39.4\%) & 4.56$\times$ & 3.77 (43.6\%) & 4.52$\times$ \\
  \multicolumn{2}{c|}{Qwen2.5-72B} & 95.53 (100.0\%) & 1.0$\times$ & 86.59 (100.0\%) & 1.0$\times$ & 8.65 (100.0\%) & 1.0$\times$ \\
  \hline
  \multirow{3}{*}{\begin{minipage}[t]{0.1\linewidth}\centering Qwen2.5\\0.5B/72B \end{minipage}}
  & SD & 95.45 (99.9\%) & 2.22$\times$ & 87.20 (100.7\%)  & 2.18$\times$ & 8.65 (100.0\%) & 1.57$\times$ \\
  & MD & 94.92 (99.4\%) & 2.35$\times$ & 84.76 (97.9\% )& 2.31$\times$ & 8.34 (96.4\%) & 1.59$\times$ \\
  & CD & 96.13 (100.6\%) & \textbf{2.47$\times$} & 84.15 (99.3\%) & \textbf{2.42$\times$} & 8.32 (96.2\%) & \textbf{1.80$\times$} \\
  \hline
  \multicolumn{2}{c|}{Qwen2.5-7B} & 91.36 (95.6\%) & 3.75$\times$ & 83.54 (96.5\%) & 3.59$\times$ & 7.79 (90.0\%) & 4.44$\times$ \\
  \hline
  \multirow{3}{*}{\begin{minipage}[t]{0.1\linewidth}\centering Qwen2.5\\7B/72B \end{minipage}}
  & SD & 95.45 (99.9\%) & 2.09$\times$ & 86.59 (100.0\%) & 2.12$\times$ & 8.65  (100.0\%) & 1.68$\times$ \\
  & MD & 95.53 (100.0\%) & 2.24$\times$ & 88.41 (102.1\%) & 2.22$\times$ & 8.41 (97.2\%) & 1.79$\times$ \\
  & CD & 96.06 (100.6\%) & \textbf{2.50$\times$} & 89.63 (103.5\%) & \textbf{2.51$\times$} & 8.53 (98.6\%) & \textbf{1.93$\times$} \\
  \hline
  \end{tabular}
  \end{adjustbox}
  \caption{Comparing verification methods for speedup ratio, based on Generic setting, where 0.5B, 7B, and 72B all represent model sizes, where the smaller is the draft model and the larger is the target model.}
  \label{tab:results_for_ratio_preserving_acc}
\end{table*}

\subsection{Discussion about EAGLE-Style model.}
\label{sec:discussion_eagle}
As shown in Table~\ref{tab:results_for_structure}, CD demonstrates good generalization on the Eagle-style framework, indicating that CD can be well adapted to both self-drafting and dynamic tree methods. However, it should be noted that the acceleration capability of CD is related to the draft length. The Eagle-style method faces challenges in extending the draft length, which limits the potential of CD.

\section{Parameters for Best Speedup}
\label{sec:parameters-for-speed}
As shown in \Cref{tab:parameters of best speedup}, we present the parameters required for the experiments in Table~\ref{tab:results_for_ratio}: threshold and draft length. These parameters were obtained using grid search to demonstrate the optimal speedup ratio for each validation method. For the other parameters, we use the default settings provided in the Generic setting in \Cref{experiments}.

\begin{table*}
  \centering
  \begin{adjustbox}{max width=0.9\textwidth}
  \begin{tabular}{lc|cc|cc|cc|cc}
  \hline
  \multirow{2}{*}{\textbf{Model}} & \multirow{2}{*}{\textbf{Verify}} & \multicolumn{2}{c|}{\textbf{GSM8K}} & \multicolumn{2}{c|}{\textbf{HumanEval}} & \multicolumn{2}{c|}{\textbf{MT-Bench}} & \multicolumn{2}{c}{\textbf{AlpacaEval}} \\
  & & Threshold & Length & Threshold & Length & Threshold & Length & Threshold & Length \\
  \hline
  \multirow{3}{*}{\begin{minipage}[t]{0.1\linewidth}\centering Qwen2.5\\0.5B/72B \end{minipage}}
  & SD & - & 6 & - & 6 & - & 6 & - & 6 \\
  & MD & 0.7 & 10 & 0.6 & 10 & 0.8 & 8 & 0.7 & 8 \\
  & CD & 6.0 & 15 & 6.0 & 20 & 4.0 & 8 & 4.0 & 8 \\
  \hline
  \multirow{3}{*}{\begin{minipage}[t]{0.1\linewidth}\centering Qwen2.5\\7B/72B \end{minipage}}
  & SD & - & 8 & - & 8 & - & 6 & - & 6 \\
  & MD & 5.0 & 25 & 5.0 & 25 & 6.0 & 20 & 6.0 & 20 \\
  & CD & 6.5 & 25 & 7.5 & 25 & 6.0 & 25 & 5.0 & 20 \\
  \hline
  \end{tabular}
  \end{adjustbox}
  \caption{The table provide the threshold and draft length for Optimal-Speedup Setting, where "Length" is draft length. }
  \label{tab:parameters of best speedup}
\end{table*}

\end{document}